\documentclass[10pt,twocolumn,letterpaper]{article}
\usepackage{cvpr}
\usepackage{times}
\usepackage{epsfig}
\usepackage{graphicx}
\usepackage{amsmath}
\usepackage{amssymb}
\usepackage{xspace}
\usepackage[pagebackref=true,breaklinks=true,letterpaper=true,colorlinks,bookmarks=false]{hyperref}

\cvprfinalcopy % *** Uncomment this line for the final submission

\def\cvprPaperID{2067} % *** Enter the CVPR Paper ID here

\ifcvprfinal\fi
\begin{document}

%%%%%%%%% TITLE
\title{Generalisation and Sharing in Triplet Convnets for Sketch based Visual Search}

\author{Tu Bui, John Collomosse\\
University of Surrey\\
Guildford, Surrey, UK\\
{\tt\small {t.bui,j.collomosse}@surrey.ac.uk}
\and
Leonardo Ribeiro,  Moacir Ponti\\
University of Sao Paulo\\
Butanta, Sao Paulo, Brazil\\
{\tt\small {l.ribeiro,ponti}@usp.br}
}

\maketitle
%\thispagestyle{empty}

%%%%%%%%% ABSTRACT
\begin{abstract}

We propose and evaluate several triplet CNN architectures for measuring the similarity between sketches and photographs, within the context of the sketch based image retrieval (SBIR) task. In contrast to recent fine-grained SBIR work, we study the ability of our networks to generalise across diverse object categories from limited training data, and explore in detail strategies for weight sharing, pre-processing, data augmentation and dimensionality reduction.  We exceed the performance of pre-existing techniques on both the Flickr15k category level SBIR benchmark by $18\%$, and the TU-Berlin SBIR benchmark by $\sim10 \mathcal{T}_b$, when trained on the 250 category TU-Berlin classification dataset augmented with 25k corresponding photographs harvested from the Internet.

\end{abstract}

%%%%%%%%% BODY TEXT
\section{Introduction}

Sketches are an intuitive and concise modality for communicating everyday concepts; abstract visual depictions pre-date the written language of our early ancestors, and drawings are one of the earliest communication mediums employed by children.  With the advent of modern touch-screen based devices, gestural forms of interaction are of increasing interest as a means for navigating visual media. This paper addresses the particular problem of {\em sketch based image retrieval (SBIR)}; searching a collection of photographs (images) for a particular visual concept using a free-hand sketched query.

The principal contribution of this paper is to explore SBIR from the perspective of a cross-domain modelling problem, in which a low dimensional embedding is learned between the space of sketches and photographs.  Historically, SBIR has been addressed using sparse feature extraction and dictionary learning approaches, following the successful application of the same to object recognition and the measurement of visual similarity in natural images \cite{Hu2010,Eitz2012,Hu2013}.  Deep convolutional neural networks (convnets/CNNs) have since gained traction as a powerful and flexible tool for tackling  a diverse range of machine perception problems \cite{Krizhevsky2012}, and very recently have also been explored for SBIR within the context of fine-grain retrieval, e.g. to find a specific shoe within a dataset of only shoes \cite{Yu2016,Sangkloy2016}.  Our work is aligned with the larger body of SBIR work addressing the problem of category-level retrieval, in which a user sketches a kind of object with particular attributes (\eg a specific furniture form, a spotted dog, or a  building with particular structure), seeking images that conform to that structure rather than a specific single image \cite{Eitz2009,Eitz2012,Hu2013,Saavedra2014,Saavedra2015}.  Acknowledging the complementarity and value in both perspectives, this paper for the first time explores deep learning for category-level SBIR;  compatible with the inherent ambiguity of sketch and the common use case of sketching an unseen prototypical object `from the mind's eye' (\eg web search), rather than to recall a previous observation of a specific object to the last detail.  Specifically this paper explores appropriate convnet architectures, weight sharing schemes and training methodologies to learn a low-dimensional embedding for the representation of both sketches and photographs --- in practical terms, a space amenable to fast approximate nearest neighbour (ANN) search (\eg $L^2$ norm) for high performance SBIR. We explore several important questions around  effective deep learning of such representations. 

1. {\bf Generalisation:} Given the diversity of visual concepts in the wild ($\sim10^5$ categories) and the challenges of  annotating large sketch datasets (current best,  $\sim10^2$ categories \cite{Eitz2012}) how well can a CNN generalise beyond its training concepts to represent unseen sketched object categories? Are class diversity and volume of exemplars equally important?

2. {\bf Input Modality:} SBIR and the related task of sketched image classification variously employ edge extraction as a pre-processing step to align the statistics of sketch and photo distributions. Is this  a beneficial strategy  when learning a SBIR feature embedding?

3. {\bf Architecture:} Recent exploration of SBIR has indicated triplet loss CNNs as a promising archetype for SBIR embedding, however what kind of loss objective should be considered and where, and which weight sharing strategies are most effective? What is the best way to enforce a low dimensional embedding for efficient SBIR indexing?

\section{Related Work}

Sketch based Image Retrieval (SBIR) began to gain momentum in the early nineties with colour-blob based query systems such as Flickner \etal's QBIC  \cite{QBIC1995} that matched coarse attributes of colour, shape and texture using region adjacency graphs.  Several global image descriptors for matching blob based queries were subsequently proposed, using spectral signatures derived from Haar Wavelets \cite{Jacobs1995} and the Short-Time Fourier Transform \cite{Sciascio1999}.  This early wave of SBIR systems was complemented in the late nineties by algorithms accepting line-art sketches, more closely resembling
the free-hand sketches casually generated by lay users in the act of sketching a throw-away query \cite{Collomosse2008}.  Such systems are characterised by their optimisation based matching approach; fitting the sketch under a deformable model to measure the support for  sketched structure within each photograph in the database \cite{Bimbo1997,Collomosse2009}.  Despite good accuracy, such approaches are slow and scale at best linearly.  It was not until comparatively recently that global image descriptors were derived from line-art sketches, enabling more scalable indexing solutions. 

\subsection{SBIR with shallow features}

Mirroring the success of gradient domain features and dictionary learning methods to photo retrieval, both Hu \etal \cite{Hu2010} and Eitz \etal \cite{Eitz2011} extended the Bag of Visual Words (BoVW) pipeline to SBIR, notably also proposing the Flickr15k dataset which became a defacto benchmark for category-level SBIR \cite{Hu2013}. Sparse features including the Structure Tensor \cite{Eitz2009}, SHoG \cite{Eitz2011}, Gradient Field Histogram of Oriented Gradients (GF-HOG) \cite{Hu2013} and its extended version~\cite{Bui2015} are extracted from images pre-processed via Canny edge detection.  Mid-level features were also explored through the HELO and key-shapes schemes of Saavedra and Bebustos \cite{Saavedra2014,Saavedra2015}, which although  not indexable via BoVW could be matched via Hungarian algorithm.  Mid-level structures were also explored in the Mindfinder system of Cao \etal \cite{Cao2011} who were the first to propose inverse index structure for scalable SBIR.  Such systems score around 10-15\% on the Flickr15k dataset often failing in the presence of edge clutter. Recently, Qi \etal employed an alternative edge detection pre-process raising the state of the art to 18\% over Flickr15k. Their Perceptual Edge \cite{Qi2015} filter simplifies the edge map prior to HoG/BoVW  delivering a performance gain in cluttered scenes.

\subsection{SBIR with deep networks}

The use of deep networks to obtain data-driven representations can potentially overcome the challenges of learning from different domains. An example of single network model is the SketchANet~\cite{Yu2015}, a smaller version of AlexNet optimised for sketch recognition only. However, while learning from a single domain is straightforward --- provided enough data --- mapping between a sketch and a photograph often requires multi-branch networks for cross-domain mapping. In this context, it is possible to learn representations for each domain independently, or perform a multi-domain learning by sharing knowledge across the different -- but related -- domains. Triplet convnets are especially interesting in this scenario since they employ three branches \cite{Wang2014}: an anchor branch, which models the reference object, one branch representing positive examples (which should be similar to the anchor) and another modeling negative examples (which should be as different as possible to the anchor). The triplet loss function is responsible to guide the training stage considering the relationship between the three models. This method was used for photographic queries by Wang \etal~\cite{Wang2014} and for sketched queries in order to refine search within a specific object class (\eg a dataset of shoes)~\cite{Yu2016}. Similarly, a fine-grained approach to SBIR was adopted by the recent Sketchy system of Sangkloy \etal \cite{Sangkloy2016} in which careful reproduction of stroke detail is invited for instance-level search.

In Yu \etal \cite{Yu2016}, the authors train one model for each target category, the sketch is matched against the edgemap extracted from a well-behaved image (without clutter, often with constant background). They report that using a fully-shared network was better than use two networks without weight sharing. However, in category-level problem it may be beneficial to avoid sharing all layers to encourage generalisation (c.f. in Sec.~\ref{sec:eval:spn}). Wang \etal~\cite{Wang2015} used a two-branch network with contrastive loss for sketch-based 3D shape retrieval without sharing weights, indicating that when two domains are very different sharing knowledge may not be ideal. Therefore, questions remain around training methodology including architecture, weight-sharing strategies, and loss functions, as well as the generalisation capability of such models.  Our work explores these questions, and broadens the investigation of deep learning to SBIR beyond intra-class or instance level search to retrieval across multiple object categories.  To avoid confusion we hereafter refer as {\em no-share} or Heterogeneous those multi-branch networks for which there are no shared weights between layers~\cite{Wang2015}; as {\em full-share} or Siamese those for which all layers have shared weights in all layers~\cite{Yu2016,Wang2014}; and {\em half-share} or Hybrid those for which only a subset of layers are shared.

%-----------------------------------------------------------------------------------------------------------------------
\section{Methodology}

We propose and investigate several triplet network designs, comparing the Siamese architecture with the Heterogeneous and a Hybrid design. Triplet networks are commonly used to learn joint embeddings from data distributions, and have been recently applied to photographic visual search \cite{gordo2016deep,CTU-ECCV2016}. In an image retrieval context, a query image  is presented to the anchor branch, and relevant/irrelevant images are presented to the positive and negative branches respectively. The positive and negative branches share their weights because they represent the same domain (\eg photos). The anchor branch represents a different domain (sketch) and differences in design arise from the degree to which layers within the anchor branch share weights with the other branches.  One or several fully-connected (FC) layers unifying the branches may be installed, as may loss functions pre- and post- unification.  In this section we explore several permutations of this design, evaluating their performance in Sec.~\ref{sec:experiments} with the aim of testing the generalisation capability of the network across categories, and identifying the best performing architecture (CNN architecture, loss) and training strategy to optimise retrieval accuracy.

\subsection{Network architecture}
\label{sec:network}
Whilst some SBIR techniques perform feature extraction directly from image pixels, several pre-process images into edge-maps \cite{Yu2016,SongBMVC16}.  We explore both strategies through two base architectures in our experiments.

\begin{figure*}
\begin{center}
    \begin{tabular}{ccc}
        \includegraphics[width=0.3\linewidth,height=5.2cm]{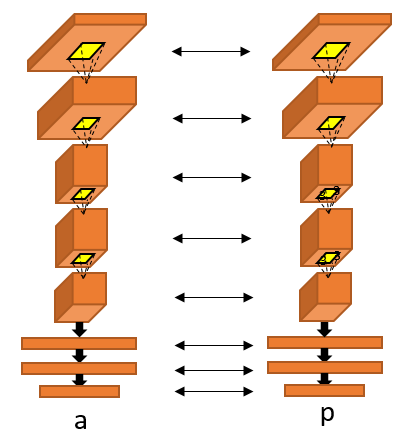} &
        \includegraphics[width=0.25\linewidth,height=5.2cm]{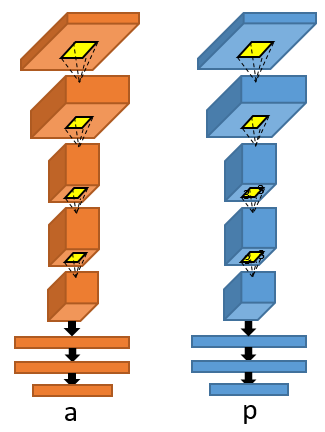} &
        \includegraphics[width=0.3\linewidth,height=5.2cm]{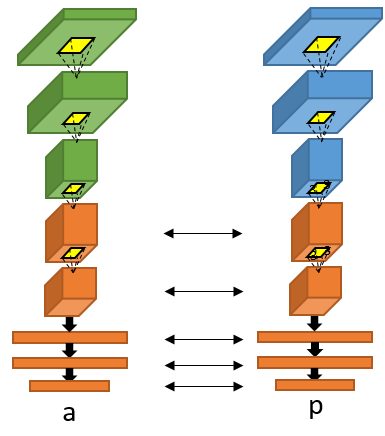} \\
        \includegraphics[width=0.3\linewidth]{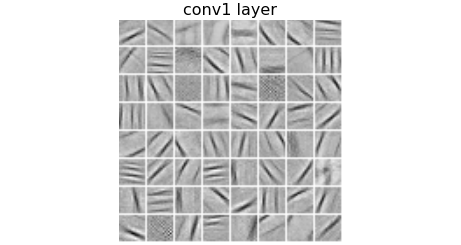} &
        \includegraphics[width=0.3\linewidth]{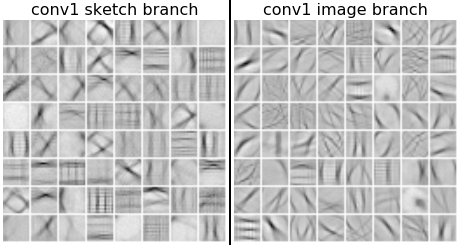} &
        \includegraphics[width=0.3\linewidth]{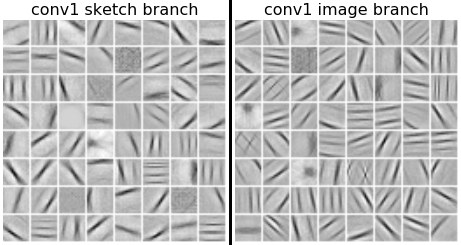} \\
    \end{tabular}
\end{center}
   \caption{Three triplet network designs evaluated for sketch-edgemap matching and the corresponding visualisation of its first convolutional layer. Arrows/same colour indicates sharing of weights. For simplicity we omit the negative branch. Note the effect of integrating classification loss into the Heterogeneous network (middle), resulting in a learned convolution layer contrasting with the Siamese (left) and half-share (right) networks.}
\label{fig:sketch-edgemap-arch}
\end{figure*}

\paragraph{Sketch-edgemap matching}: we utilise the SketchANet architecture~\cite{Yu2015} for both sketch and edgemap (positive/negative) branches, adopting similar branch design on the assumption that a photo's edgemap is statistically closer to a sketch. The SketchANet network is similar to AlexNet \cite{Krizhevsky2012} but 8 times smaller and optimised for sketch, which is ideal for applications that require fast processing time like SBIR. Fig.~\ref{fig:sketch-edgemap-arch} depicts three weight sharing variants explored in this work. The full-share network (Siamese) has three identical branches. The no-share network (Heterogeneous) allows the anchor branch to be learned independently. The half-share network (Hybrid) has part of the anchor branch shared with the two others. Since the top layers of a CNN network learn high-level concepts we shared the top 4 layers among the 3 branches. %Note that although we do not evaluate all half-share layer permutations we later observe that best results arise under this design for the Siamese variant.

\begin{figure}[t]
\begin{center}
    \begin{tabular}{c}
        \includegraphics[width=0.9\linewidth]{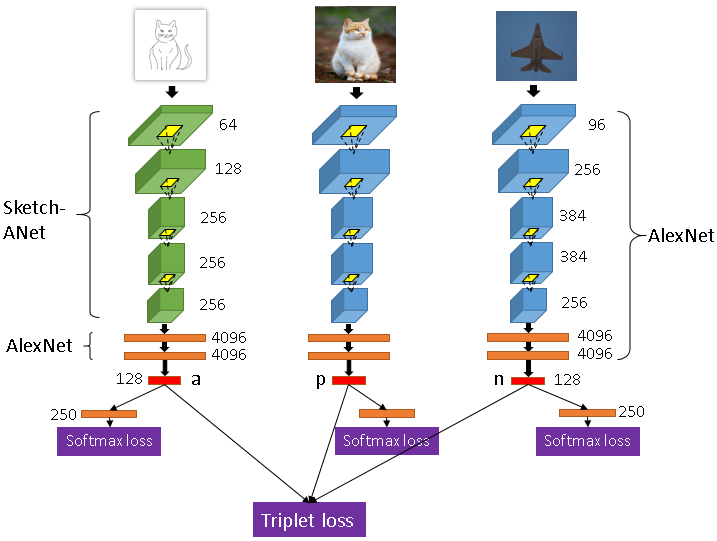}\\
        \includegraphics[width=0.75\linewidth]{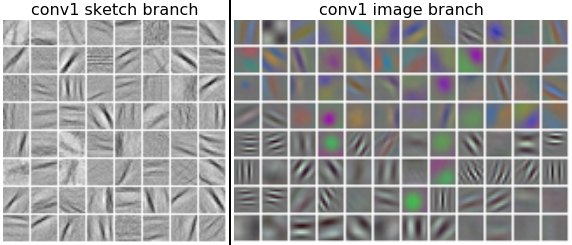}\\
    \end{tabular}
\end{center}
  \caption{Half-share triplet network for sketch-photo direct mapping hybridised from the AlexNet and SketchANet designs. Layers with the same colour indicate weight sharing.The dimensionality reduction layer $FC_R$(red) is not associated with any activation function. The 4 FC layers can alternatively be configured unshared for a Heterogeneous network. When training this network the softmax loss layer is only activated during the training phase 2.}
\label{fig:sketch-photo-arch}
\end{figure}

\paragraph{Sketch-photo matching}: we utilise the AlexNet CNN for the photo branches, and a hybrid CNN combining AlexNet and SketchANet for the sketch branch that shares layers 6-7 between the sketch and photo branches (see Fig.~\ref{fig:sketch-photo-arch}). Specifically, layers 1-3 have SketchANet architecture, layers 6-7 mirror AlexNet while the middle layers 4-5 we have modified from SketchANet as a hybridisation of the two designs. We configure fewer shareable layers than sketch-edgemap's because the domain gap between sketch and photo is larger. Nevertheless, layers 6 and 7 alone account for more than 90\% of the parameters of both networks. To the best of our knowledge, this is the first triplet network presented for visual search that has different architecture for the anchor and the positive/negative branches.

\subsection{Dimensionality reduction}
It is desirable in visual search to have a compact representation of images. Dimensionality reduction is generally within a CNN by adding an intermediate fully-connected (FC) layer with a lower number of nodes. In a classification network, this number is constrained by number of categories i.e. intermediate FC layer should be larger than the final $D$-way FC layer where D is class number. An advantage of the triplet network is that the dimensionality of the output embedding can be freely set, since the triplet loss function only compares the output features of the three branches, and does not take into account labels of the input. In all experiments we fix the embedding dimension $D=128$, regardless of network design or number of categories in the training set. This allows us to encode the whole Flickr15K dataset using just 7MB memory.

Even when classification loss must be integrated in the network during certain training phases (see Sec.~\ref{sub:train_pro}), we propose an unique way to overcome the problem as illustrated in Fig.~\ref{fig:sketch-photo-arch}. Here we add an embedding layer $FC_R$ ($D=128$) between layer FC7 ($D=4096$) and FC8 ($D=250$) without being followed by an activation (ReLU) layer. This prevents the embedding layer becoming a bottleneck in the network, since from the perspective of the softmax-loss layer the connection from FC7 to FC8 is  linear. We empirically verify that during training the performance of the classification layer is not affected whether $FC_R$ is integrated or not.

\subsection{Training procedure}
\label{sub:train_pro}

We now describe the different training strategies for different network designs. Practically we observed that  difficulties in achieving convergence increase from full-share, half-share to no-share, and that due to the domain gap, learning a sketch-photo mapping is more challenging that the sketch-edgemap case. In all experiments, we steer away from using classification loss to prevent affecting the generalisation capability of the network. Nevertheless we found that integrating softmax-loss, even if only at an early phase, is unavoidable for the training to converge in some cases.

\noindent-- \textbf{Full-share sketch-edgemap}: this configuration was trainable using standard triplet loss in a single step.\\
-- \textbf{Half-share sketch-edgemap}: this configuration encounters the so called ``gradient vanishing problem". We proposed a new loss function to overcome it (Sec.~\ref{sub:grad_van}).\\
-- \textbf{No-share sketch-edgemap}: we conduct a two-step training process: (i) pre-train the sketch and edgemap branches separately using softmax-loss; (ii) remove softmax-loss and train the network further with the triplet-loss.

Given the instantiation of existing networks  with branches of the triplet CNN, an opportunity exists to exploit {\ a prior} trained models.  Although we leverage fine-tuning in our experiments training sketch-photo networks, (subsec.~\ref{sec:eval:spn})  we train all networks from scratch during our generalisation experiments (subsec.~\ref{sec:eval:gen}).

Sketch-photo \ie cross-domain learning requires more  effort to achieve good convergence. We use the following multi-step training strategy:\\
-- \textbf{Phase 1: Classification} train the sketch and photo branches independently using a softmax-loss at FC8.\\
-- \textbf{Phase 2: Classification + regression} for the half-share network only. We form a double-branch network, freeze the unshared layers which were already learnt during phase 1. Next, we use contrastive loss together with softmax-loss to train the sharing layers. The use of softmax-loss in such double-branch network helps the sharing layers to learn discriminative features from both sketch and edgemap domains, whilst contrastive-loss provides an early step of regression to bring the two domains together.\\
-- \textbf{Phase 3: Regression} Remove softmax-loss and its associated fully-connected layer, unfreeze all frozen layers and train the network with triplet loss.\\
-- \textbf{Phase 4 (optional)} Any auxiliary sketch-photo datasets available can be used to further refine the model.

In the above training procedure, classification loss is used to guide the training during the early phases. Without it the training would not converge. The latter phases  purely use regression to deliver the embedding for SBIR. 

\paragraph{Data augmentation} plays an important role in preventing overfitting, especially when training data is limited. In all experiments we apply the following augmentation techniques for both sketch/edgemap and photo: random crop (sketch/edgemap crop size 225x225 for SketchANet, photo crop size 224x224 for Alexnet), random rotation in range $[-5,5]$ degrees, random scaling in range $[0.9,1.1]$ and random horizontal flip.  

We also propose an augmentation method applicable for sketches only.  For sketches with at least $N$ strokes ($N=10$ in our experiments) we divide them into four equal groups of strokes in drawing order. The first group contains the most important strokes --- related to the more coarse structure of the object --- and it is always kept. A new sketch is created by randomly discarding some of the other groups. This technique is inspired by Yu \etal \cite{Yu2015,Yu2016} who observe that people tend to draw sketches in stages at distinct levels of  abstraction. We observed a $\sim1\%$ mAP improvement across the board using this random stroke removal augmentation method on the Flickr15k benchmark.

\subsection{The gradient vanishing problem}
\label{sub:grad_van}
\begin{figure*}
\begin{center}
    \begin{tabular}{cccc}
           \includegraphics[width=0.36\linewidth]{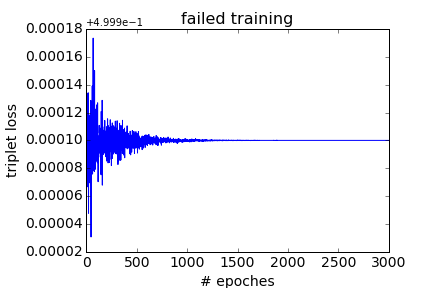} &
           \includegraphics[width=0.15\linewidth]{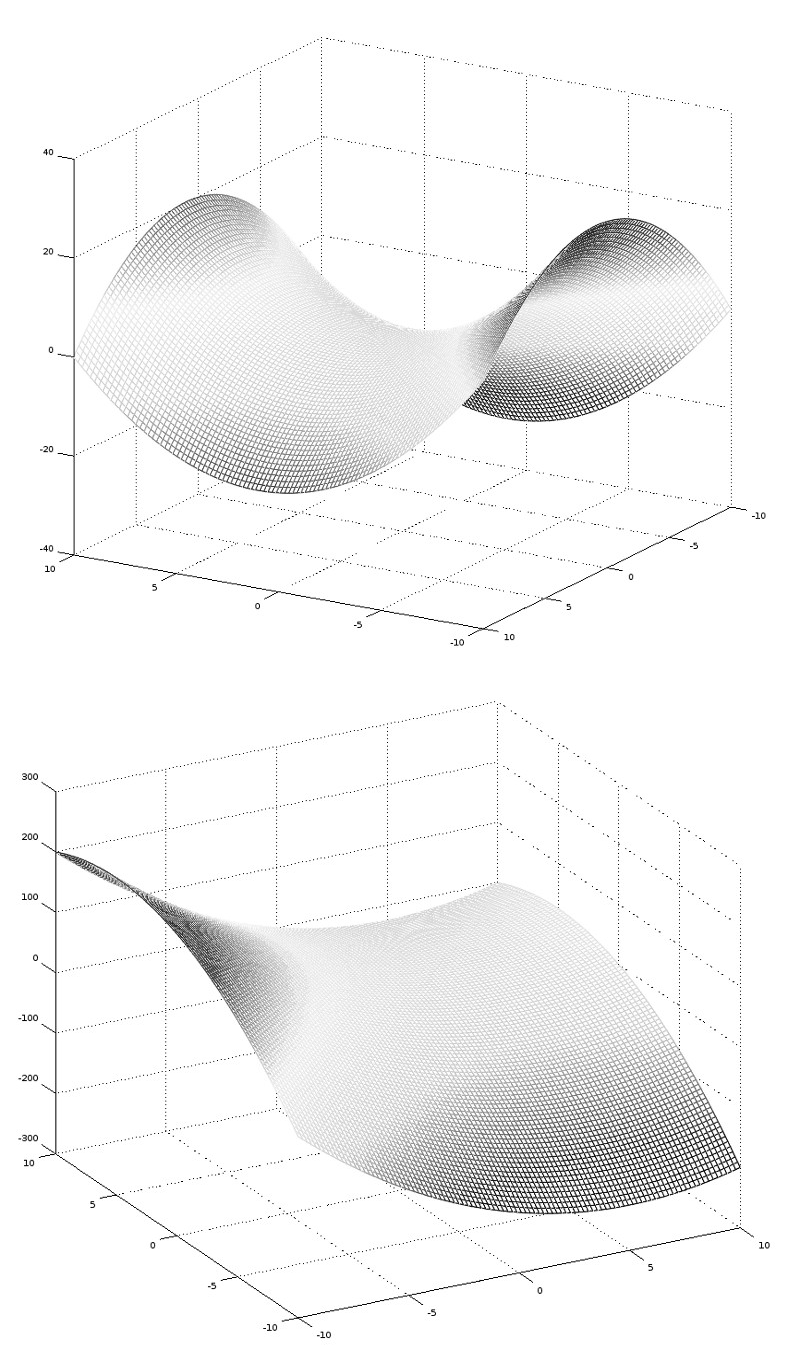} &
           \includegraphics[width=0.36\linewidth]{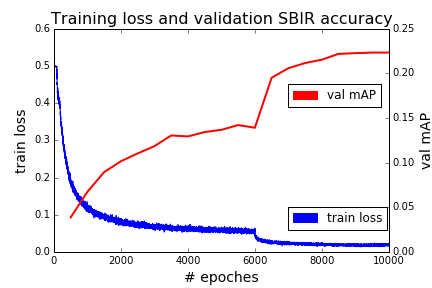} \\
     \end{tabular}
\end{center}
   \caption{The gradient vanishing problem: (left) a training failure case where loss flattens at half of the margin; (middle top) a saddle created in the loss space when anchor, positive and negative points are very close to each other; (middle bottom) modified loss slope ideal for gradient descent when anchor point is far from positive and negative points; (right) successful training of the half-share network with new loss function \ref{eq:newlossfunction}. Note: the validation mAP is the retrieval score when querying the validation sketch set against the training photos.}
\label{fig:grad_van}
\end{figure*}

The gradient vanishing problem \cite{Wang2014} manifests itself during the training of several of the above network designs where the vectors output from the three network branches fall very close to each other. In this scenario, instead of pushing the negative feature point away while bringing the positive point closer to the anchor point, the training ends up equalising distance between all three points (Fig.~\ref{fig:grad_van}-left). Consequently the standard triplet loss flattens at half of the margin and stops learning (its derivative w.r.t each input becomes zero). We hypothesise that the training probably falls to a ``saddle'' space and unable to move down the either way (illustrated in Fig.~\ref{fig:grad_van}-middle). We adopt a modified loss function to prevent ``saddle'' from being created at the outset of  training: 
\begin{equation}
    \label{eq:newlossfunction}
   \mathcal{L}(a,p,n) = \frac{1}{2N} \sum_N \max \left[ 0, m + |2a - p|^2 - |2a-n|^2 \right]
\end{equation}
where $a$, $p$ and $n$ are output features from the anchor, positive and negative branches respectively; $m$ is the margin and $N$ is size of the training mini-batch.

Under this loss, forcing sketch feature to be relatively half of the edgemap feature will break that balance, thus creating a ``slope" for gradient descent to converge (Fig.~\ref{fig:grad_van}-right). At query time, the sketch (query) feature is scaled by 2.0 to match the scale of features extracted from the photos.

%---------------------------------------------------------------------------------------------------------------
\section{Experiments}
\label{sec:experiments}

We now evaluate the variants of our proposed triplet architecture and weight sharing schemes to determine best performance embedding for SBIR.  In particular we evaluate the ability of the network to generalise beyond the categories to which it is exposed during training.  This is important for SBIR in the wild, where one cannot reasonably train with a sufficiently diverse sample of potential query images.  We also investigate the impact of volume of sketch data used to train the network, and the impact of using photos or their edge-maps during training (in addition to the various weight sharing variants).

\subsection{Datasets}
\label{sec:datasets}

We train and evaluate our networks using four sketch datasets: the TU-Berlin datasets for classification~\cite{Eitz2012} and SBIR~\cite{Eitz2011}, the Flickr15K benchmark dataset for SBIR~\cite{Hu2013}, and the Sketchy~\cite{Sangkloy2016} dataset:

-- {\bf TU-Berlin-Class}~\cite{Eitz2012} (used here for training) for sketch classification comprising 250 categories of sketches, 80 per category, crowd-sourced from 1350 different non-expert participants with diverse drawing styles; 

% Leo: this description doesn't specify the number of sketches and it is not very clear the dataset is composed of sketches and rankings. I propose the followiing change:
% TU-Berlin ...cite{...} is composed of 31 sketches and 40 corresponding images and ranks of those for each sketch (1240 total images). The authors propose the use of Kendall's tau rank correlation measure as the evaluation method; this is a test that assumes the user searches for images of a target object, but considering a desired pose and spatial arrangement;
-- {\bf TU-Berlin-Retr}~\cite{Eitz2011} (used here for testing) takes into account not only the category of the retrieved images but also the relative order of the relevant images. The dataset consists of 31 sketches and 40 ranked images for each sketch (1240 total images), mixed with a set of 100,000 distracting Flickr photos. The authors propose a Kendal score as the evaluation method;

-- {\bf Sketchy}~\cite{Sangkloy2016} (used here for model fine-tuning) is a recent fine-grained dataset in which each photo image has $\sim5$ instance-level matching sketches drawn by different subjects.  The dataset has only 125 categories but over 500 sketches per category;  

-- {\bf Flickr15K}~\cite{Hu2013} (used for testing) has labels for 33 categories of sketches, 10 sketches per category. It also has a different number of photo images per category totalling 15,024 images. The Flickr15K has only four categories that are similar to the ones in TU-Berlin-Class dataset: ``swan'', ``flowers'', ``bicycle'' and ``airplane''. Those two datasets also differ in the concept of some categories, for example TU-Berlin has a general ``bridge'' category while Flickr15K distinguishes ``London bridge'', ``Oxford bridge'' and ``Sydney bridge''.  Also, their sketch depiction is different, motivating a need for good generalisation beyond training.

As TU-Berlin-Class comprises only sketches, in order to obtain our training triplets we automatically generate per-category photograph sets by querying the 250 category names on Creative Commons image repositories. The Flickr API was used to download images from 184 categories. Google and Bing engines were used for the remaining 66 categories which are mainly human body parts (e.g. brain, tooth, skeleton) and fictional objects (e.g. UFO, mermaid, dragon) where Flickr content was sparse. We manually selected the 100 most relevant photos per category to form our training data.

\subsection{Experimental settings}
We use the TU-Berlin-Class as the training sketch set.  Sketches are skeletonised to be consistent with the photo's edgemap.  For the sketch-edgemap experiments, we extracted the edgemaps of the photo dataset using a state-of-the-art non-deep learning detector, gPb~\cite{Arbelaez2011}, after \cite{Qi2015}.  Hysteresis thresholding is applied on the resulting gPb soft-edge as per Canny edge detection to remove weak edges as well as isolated edge pixels.

As test set, we use the category-level SBIR Flicrk15K benchmark~\cite{Hu2013}. Additionally, we also tested our models on the TU-Berlin-Retr~\cite{Eitz2011}. Note that this dataset is completely different from  TU-Berlin-Class~\cite{Eitz2012} (subsec.~\ref{sec:datasets}). While the performance metric for the Flickr15K benchmark is retrieval Mean Average Precision (mAP), the TU-Berlin-Retr benchmark supports labels for the order of relevance of the returned photos so the Kendall's rank correlation coefficient was employed. 

A series of experiments was carried out, starting with a subset of 20 random training categories and 20 sketches per category, up to the whole training dataset. As the TU-Berlin-Class has 80 sketches per category, the remaining sketches of the chosen categories are used for validation. We use Caffe the deep-learning library~\cite{jia2014caffe} for the training tasks. When training the triplet network, positive and negative photo samples are selected randomly. 
%We also tried hard negative mining but the process often ends up selecting outliers as negative examples, thus degrades the generation performance of the network. 
%\emph{(\color{red}TU: hey, it is actually a hole here. I tested hard-neg mining for the sketch-edgemap only. Due to bad edge extraction it causes outliers being selected as neg samples. But with sketch-photo mapping, we are having a very clean photo dataset so I believe hard neg mining will do a better job - just don't have time to prove it)}.
% Moacir: maybe we should remove the last sentence then. I commented it out, just in case.

\subsection{Generalisation and weight sharing}
\label{sec:eval:gen}
\begin{figure}[t]
\begin{center}
\begin{tabular}{c}
  \includegraphics[width=0.78\linewidth]{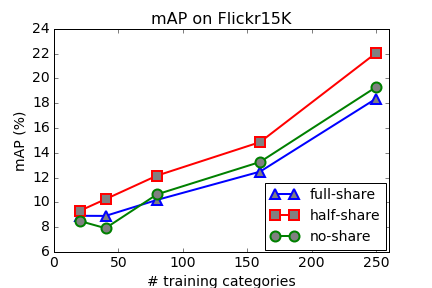}  \\
  \includegraphics[width=0.78\linewidth]{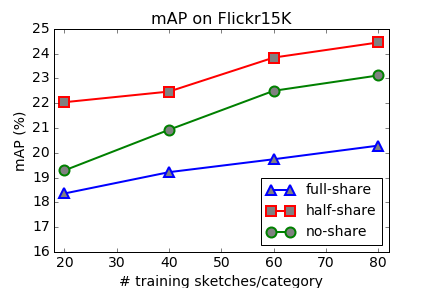}   
\end{tabular}
\end{center}
  \caption{Top: performance of the 3 different weight sharing (sketch-edgemap) strategies on Flickr15K, with the half-sharing scheme outperforming the others. Number of training sketches per category: 20. Bottom: generalisation with respect to number of training sketches per category using 250 training categories.}
\label{fig:gen2}
\end{figure}

%TODO: We can only test the generalisation for sketch-edgemap matching.
We first report the results of generalisation capability of our sketch-edgemap triplet network with varying amount of training data. Fig.~\ref{fig:gen2} (top) shows that the performance is benefited by increasing the number of training categories. All three network designs achieve near-linear improvement of retrieval performance against Flickr15k benchmark (discarding the four intersecting categories with the training set) with exposure to more diverse category set during training.  The mAP of all models jumps from $\sim9\%$ to $18$-$22\%$ when raising training data from 20 to 250 categories.

% \begin{figure}[t]
% \begin{center}
%   \includegraphics[width=0.8\linewidth]{figs/mAP_vs_cats_3.png}
% \end{center}
%   \caption{Performance of the 3 different weight sharing (sketch-edgemap) strategies on the Flickr15K benchmark, with the half-sharing scheme outperforming the others. Number of training sketches per category: 20.}
% \label{fig:generalisation}
% \end{figure}

% \begin{figure}[t]
% \begin{center}
%   \includegraphics[width=0.8\linewidth]{figs/training_loss_mAP_vs_nsketches.png}
% \end{center}
%   \caption{Network generalisation with respect to number of training sketches per category. Number of training categories: 250.}
% \label{fig:gen2}
% \end{figure}

Fig.~\ref{fig:gen2} (top) also indicates the superior performance of the half-share triplet architecture against the others in sketch-edgemap matching. Although all three perform the same with 20-categories training data, the half-share model outperforms the alternatives by $\sim 3\%$ mAP when more categories are available. Additionally, the no-share network, despite being pretrained (via step 1 of our training process, c.f. subsec~\ref{sec:network}) with softmax loss, outperforms the Siamese configuration by $\sim 1\%$ mAP. This implies that (i) sketch and photo's edgemap, although assumed statistically similar, should be treated as two different domains; (ii) sharing the top layers of the two branches can deliver a significant improvement when the two input domains are similar.

Next, we test the performance with different number of sketches per category during training. To do so we fix the number of training categories at 250, and vary the number of training sketches per category from 20 to 80 samples while keeping the other settings. All models benefit from using more sketches, with a more notable boost for half and no-share models. In Fig.~\ref{fig:gen2} (bottom) we observe an improvement of 2.5\% on the half-share model, which is significant given that with 250 training categories this model had already achieved a high mAP.

% Leo: when mentioning less layer sharing here, I think it is not clear enough this is a comparison between this and sketch-edgemap half-share. I suggest adding something like "(..) Here (as opposed to the half-share architecture on sketch-edgemap)"
\subsection{Modality: Sketch-photo vs. Sketch-edgemap}
\label{sec:eval:spn}
We tested two network designs as depicted in Fig.~\ref{fig:sketch-photo-arch}: one sharing all FC layers (half-share), and another without sharing any layers (no-share).  The training procedure for the two networks differs only during the second phase as described in Sec.~\ref{sub:train_pro}.  Table \ref{tab:flickr15k_benchmark} reports the mAP scores when compared with the best score of the sketch-edgemap network and other baselines on the Flickr15K benchmark.  Observe that: (i) All the deep-learning methods outperform the traditional methods using shallow-features by a large margin. (ii) There is also a dramatic improvement in performance of the sketch-photo networks over  sketch-edgemap configurations. (iii) In the sketch-photo case, the no-share network now beats  the half-share by $\sim1\%$ mAP.  Our third observation contrasts with the results of the sketch-edgemap evaluations (Fig.~\ref{fig:gen2}) where the half-share network performs more strongly. 
One explanation may be the significant difference between sketch and photo domains; the high-level concepts represented by the higher layers might not coincide (e.g. a sketched cat is different from a cat photo) due to abstraction, or caricaturing in the sketch itself.

\begin{table}
\begin{center}
\begin{tabular}{|l|c|}
\hline
Method & mAP (\%) \\
\hline\hline
\textbf{Triplet (fine-tuned final model)} & \textbf{36.17} \\
\textbf{Triplet (sketch-photo, no-share)} & \textbf{32.87} \\
\textbf{Triplet (sketch-photo, half-share)} & \textbf{31.38} \\
\textbf{Triplet (sketch-edgem, half-share)} & \textbf{24.45} \\
Perceptual Edge~\cite{Qi2015} & 18.37 \\
Extended GF-HoG~\cite{Bui2015} & 18.20 \\
GF-HoG~\cite{Hu2013} & 12.22 \\
SHoG~\cite{Eitz2011} & 10.93 \\
SSIM~\cite{Shechtman2007} & 9.57 \\
SIFT~\cite{Lowe2004} & 9.11 \\
Shape Context~\cite{Belongie2002} & 8.14 \\
Structure Tensor~\cite{Eitz2009} & 7.98 \\
\hline
\end{tabular}
\end{center}
\caption{SBIR comparison results (mAP) on the Flickr15K benchmark. The final model is achieved by fine-tuning the best model (sketch-photo, no-share) over the Sketchy dataset.}
\label{tab:flickr15k_benchmark}
\end{table}

\paragraph{Fine-tuned final model}: our learned models so far were trained through the first 3 phases outlined in subsec.~\ref{sub:train_pro}, in which the training sketches and photos from TU-Berlin-Class have only class labels. In phase 4, we create the final model by fine-tuning our best trained model (i.e. sketch-photo, no-share triplet) using the recently released Sketchy dataset. Note that we do not require alignment between the category sets of Sketchy and TU-Berlin-Class since we perform regression using the triplet (not classification) loss only at this stage.  This 4th stage adds fine-grained level of regression: each epoch is a pass-through of the training photos, for each photo image entering the positive branch we choose a random instance-level matching sketch entering the anchor branch and a random photo of the same category for the negative branch. The use of the Sketchy data delivers a final boost to our best case model of 3\% mAP on the standard Flickr15K benchmark (Fig.~\ref{tab:flickr15k_benchmark}). 

\subsection{Performance over SBIR benchmarks}

\begin{figure}[t!]
\begin{center}
   \includegraphics[width=0.75\linewidth]{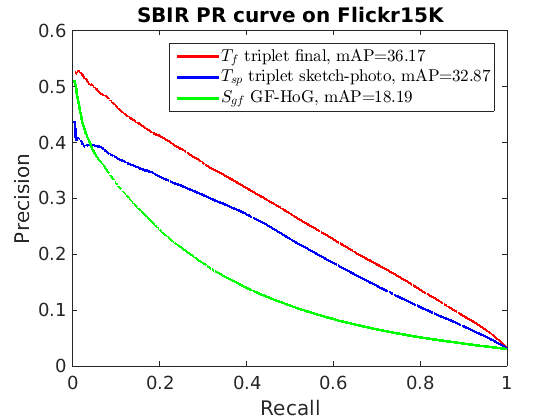}
\end{center}
   \caption{PR curve of the proposed triplet CNN (sketch-photo, c.f. Table~\ref{tab:flickr15k_benchmark}) compared with a state-of-the-art non-deep learning method \cite{Hu2013}. }
\label{fig:pr_curve}
\end{figure}

It appears that the fine-grain labels of the Sketchy dataset (in contrast to the class-level labels already provided by the training set) allow our model to learn deeper representation of photos and sketches as well as their cross-domain mapping. The improvement can be seen clearer in Fig.~\ref{fig:pr_curve}, where we plot the Precision/Recall (PR) curves of the sketch-photo no-share triplet model (snapshot after phase 3 training, denoted as $T_{sp}$), the final model ($T_{sp}$ finetuned with the Sketchy dataset, denoted as $T_f$) and one of the state-of-the-art non-deeplearning methods whose implementation code publicly available: extended GF-HoG~\cite{Bui2015} or $S_{gf}$. While the triplet sketch-photo no-share has higher mAP (blue line, mAP 32.87\%) than the extended GF-HoG (green line, mAP 18.20\%), the precision score of the first few retrieval results is actually lower for $T_{sp}$. After finetuning with the Sketchy dataset, our final model $T_f$ is able to surpass the precision score of the $S_{gf}$ at every recall point.

% Leo: There is a small typo on this figure: recal->recall

\begin{figure*}[t!]
\begin{center}
\setlength\tabcolsep{1pt}
    \begin{tabular}{cc}
           \includegraphics[width=0.45\linewidth]{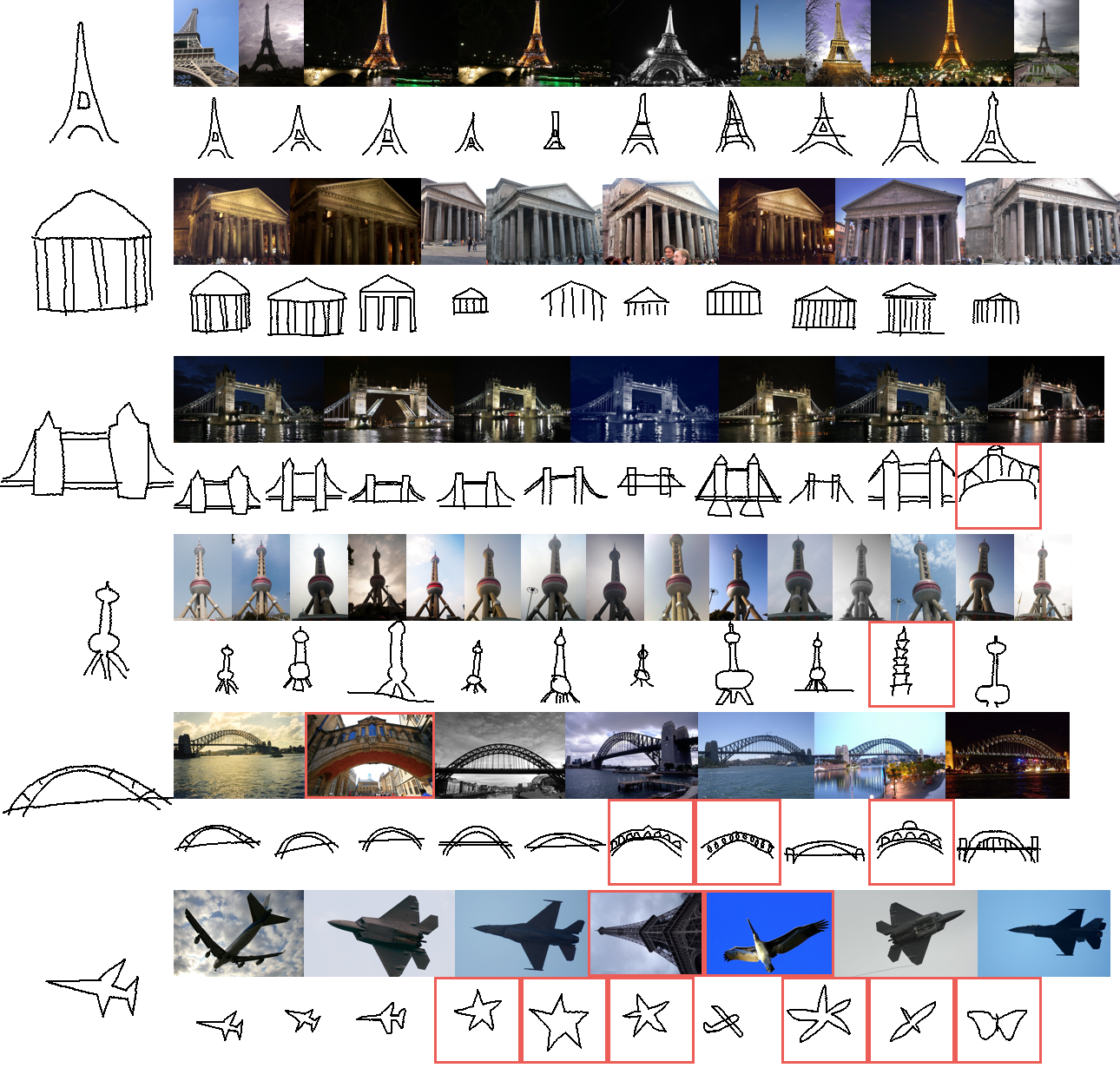} &
           \includegraphics[width=0.46\linewidth]{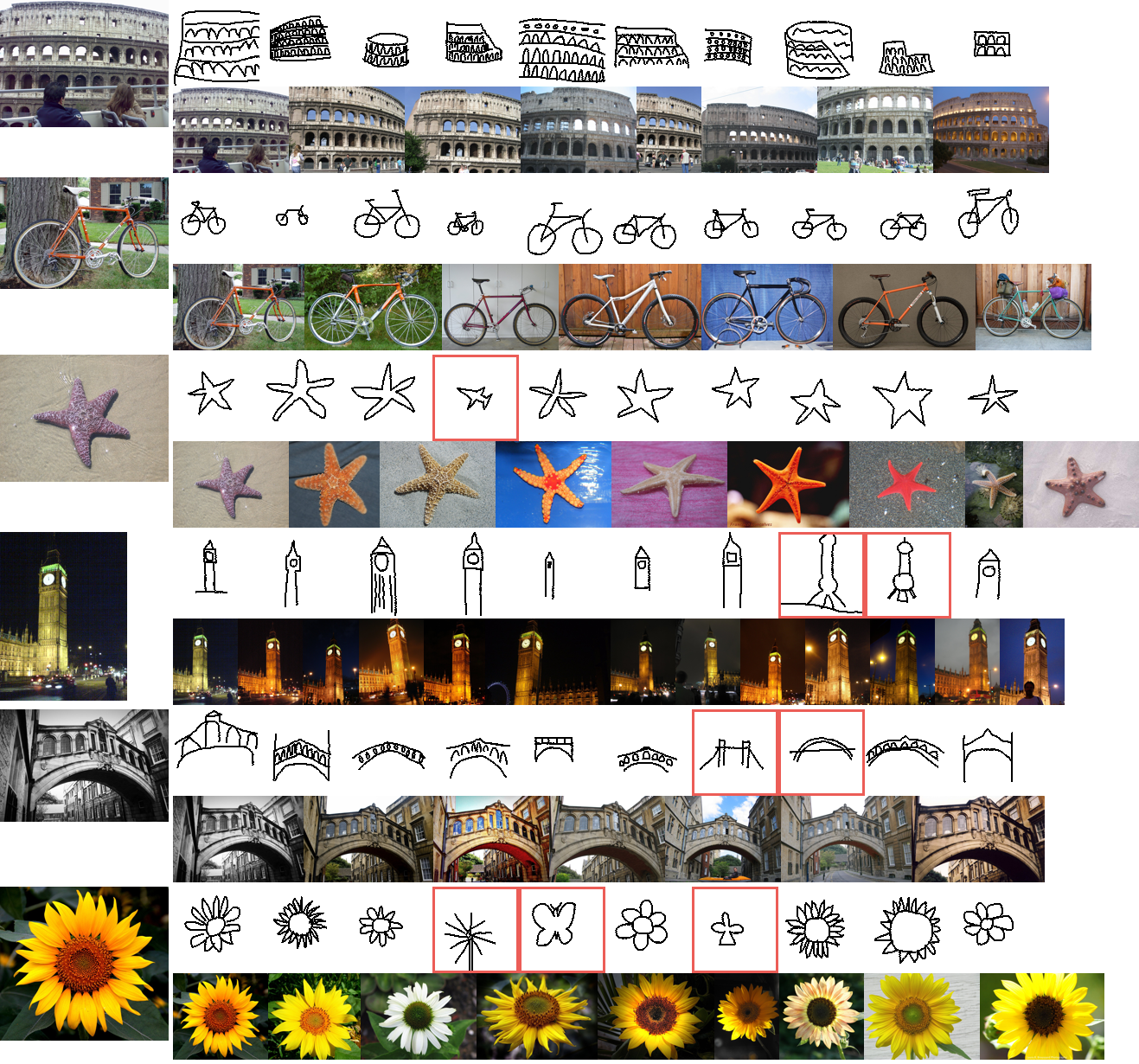} \\
     \end{tabular}
\end{center}
   \caption{Representative SBIR results on Flickr15K using (left) sketches and (right) images as queries. For each query, two sets of results are returned, one for intra-domain and the other for cross domain search. Non-relevant retrieved objects are outlined in red.}
\label{fig:SBIR_eg}
\end{figure*}

\begin{table}[t!]
\begin{center}
\begin{tabular}{|l|c|}
\hline
Method & $\mathcal{T}_b$ \\
\hline\hline
\textbf{Triplet (fine-tuned final model)} & \textbf{37.4} \\
\textbf{Triplet (sketch-photo)} & \textbf{33.3} \\
SHoG~\cite{Eitz2011} & 27.7 \\
\textbf{Triplet (sketch-edgemap)} & \textbf{22.3} \\
HoG (global)~\cite{Eitz2011} & 22.3 \\
Structure Tensor~\cite{Eitz2009} & 22.3 \\
Spark~\cite{Eitz2011} & 21.7\\
HoG (local)~\cite{Dalal2005} & 17.5\\
Shape Context~\cite{Belongie2002} & 16.1\\
\hline
\end{tabular}
\end{center}
\caption{SBIR comparison results (using Kendal's rank correlation coefficient, $\mathcal{T}_b$) on TU-Berlin-Retr dataset~\cite{Eitz2011}. The triplet sketch-edgemap model has much lower score, likely due to poor gPb edge extraction (dataset is more cluttered than Flickr15K).}
\label{tab:tuberlin_benchmark}
\end{table}

We also evaluated over TU-Berlin-Retr, using the  metric $\mathcal{T}_b$ (Table~\ref{tab:tuberlin_benchmark}) as proposed by~\cite{Eitz2011}. Interestingly the sketch-edgemap underperforms their proposed method SHoG, according to the new metric. We believe this is because TU-Berlin-Retr images are noisier than the ones on Flickr15K, which may have negative effects on the gPb edge extraction. Nevertheless, the final sketch-photo models outperform the baseline method SHoG by a significant margin.

\section{Conclusion}

We described the first deep learning algorithm for category level SBIR, comprising a triplet convnet trained using a query sketch anchor accompanied by positive and negative training photos harvested from the web.  We presented comprehensive experiments exploring variants of our triplet convnet, contrasting appropriate strategies for weight sharing, dimensionality reduction, and training data pre-processing and reporting on the generalisation capabilities of the network. We reported the half-sharing triplet performs the best when learning matching between two close domains such as sketch-edgemap, while the no-share network is more suitable for sketch-photo learning. Our best performing variant yielded a performance of 36.2\% mAP over the Flickr15k dataset and $37.4 \mathcal{T}_b$ on the TU-Berlin dataset; the twin international benchmarks for category level SBIR \cite{Eitz2011,Hu2013}. These scores exceed the state of the art by $\sim 20\%$ mAP and $\sim 10 \mathcal{T}_b$ respectively.  Training sketches were derived from the two largest available sketch datasets: the TU-Berlin dataset of Eitz \etal and the Sketchy dataset of Sangkloy \etal \cite{Sangkloy2016}. Further work might build upon this significant performance gain exploring multi-domain learning, for example sketch-photo-3D models mapping or multi-style work-art retrieval. 

\subsection{Limitations and Future Work}
The limited class diversity and volume of sketch databases prevents us from exploring our generalisation test further than the most diverse \ie 250 category dataset available (TU-Berlin-Class). Also, because of the unbalanced sketch and photo sets the sketch branch seems to be less discriminative than the photo branch. Fig.~\ref{fig:SBIR_eg} depicts several intra- and cross-domain visual search examples where false positives often occur on sketch side. Unfortunately it is difficult to control the generalisation, discrimination and overfitting of an individual branch during training of the triplet network during regression of our  SBIR embedding. 

Finally, although deep learning method proves far better than the methods using shallow features, it comes with its own cost. It is a supervised learning method which require expensive labelled training data, as opposed to a free shallow-feature extractor. A deep network that can learn cross-domain mapping with minimal or no supervision is highly desirable, especially given the content diversity faced by SBIR ``in the wild''.

%Further work might build upon this significant performance gain exploring the complementary problem of hard positive/negative mining during triplet selection.

%-------------------------------------------------------------------

{\small
\bibliographystyle{ieee}
\bibliography{egbib}
}

\end{document}